\documentclass[letterpaper, 10 pt, conference]{ieeeconf}  

\IEEEoverridecommandlockouts                              

\title{\LARGE \bf
GazeBehavior Annotation Toolkit (GBAT): AI-powered toolkit for automatic annotation of egocentric eye-tracking and video data of child-caregiver interaction
}

\author{Iba Baig$^{1,2 \dagger}$, Kevin Li$^{1 \dagger}$, Yanbin Xu$^{1 \dagger}$, Seiji Cattelain$^{3}$, Marie Hallo$^{3}$, Hayato Ono$^{4}$, Sho Tsuji$^{3,4}$ and Ming Bo Cai$^{1,4 *}$
\thanks{This work was supported by JSPS WPI-IRCN startup fund and University of Miami startup budget to MBC}
\thanks{$^{\dagger}$equal contribution}%
\thanks{$^{1}$Department of Psychology, University of Miami, $^{2}$Northeastern University, $^{3}$Ecole Normale Supérieure, PSL University, EHESS, CNRS $^{4}$International Research Center for Neurointelligence (WPI-IRCN), The
University of Tokyo Institutes for Advanced Study}%
\thanks{$^{*}$correspondence can be addressed to mingbo.cai@miami.edu}%
}
\usepackage{graphicx}
\usepackage{cite}

\begin{document}

\maketitle
\thispagestyle{empty}
\pagestyle{empty}

\begin{abstract}

Video recordings of child-caregiver interactions enable investigation of attentional dynamics during naturalistic behavior. Such multimodal recording also allows researchers to examine how attention interacts with action and language use in real time. However, manual annotation of such data is time-consuming. Here, we introduce GazeBehavior Annotation Toolkit, a deep-learning-based toolkit designed to facilitate three key processes in data preprocessing and feature extraction: post-hoc synchronization across multiple videos, semi-automatic annotation of gaze target categories, and categorization of participants’ poses and hand actions. This toolkit improves the efficiency and scalability of feature extraction from human egocentric eye-tracking and video data. Such improvement is critical in supporting large-scale and longitudinal investigations of attentional dynamics and naturalistic behavior in human early development.
\end{abstract}


\section{INTRODUCTION}
Children engage in a variety of activities driven by their intrinsic motivation to discover the underlying causal structures of the world\cite{gopnik2012scientific}. Such activities can be recorded by third-person-view cameras. During these activities, they actively sample information from the environment to guide learning through behaviors such as gaze shift \cite{yu2012embodied,smith2018developing,gureckis2012self}. In social contexts, gaze also serves a communicative function, as children express their intentions to caregivers or peers through eye movements \cite{carpenter1998social}. Head-mounted cameras or eye-trackers have been adopted to study the statistics of infants' visual input and the dynamics of attention in both children and caregivers \cite{franchak2011head,suarezrivera2019multimodal,yu2017hand,fausey2016faces}. Simultaneous recording of both the first-person and third-person view videos allows investigation of the interaction between gaze target, speech input, and body activity. 

\begin{figure}[t]
    \centering
    \includegraphics[width=\linewidth]{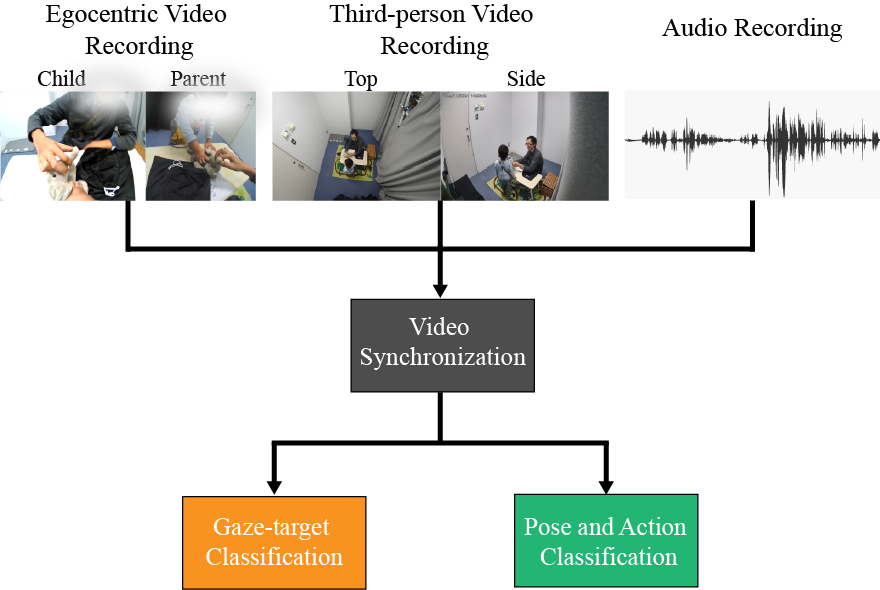}
    \caption{GazeBehavior Annotation Toolkit comes with three features: a Video Synchronizer, a Gaze Target Annotator, and a Video Content Annotator. Videos from both egocentric and third-person-view cameras are first temporally synchronized based on audio recordings. Then, gaze-target classification and pose and action classification are done using egocentric video recordings and third-person-view video recording, respectively.}
    \label{fig:overview}
\end{figure}

However, such data is of high density: children aged five to sixteen years make three to five saccades per second and videos are typically recorded at tens of frames per second\cite{larsen2017frequency}. Therefore, manual annotation of the gaze targets and the activity in the videos is labor-intensive. To facilitate the processing of multimodal recordings incorporating egocentric and third-person videos, we developed the GazeBehavior Annotation Toolkit (GBAT), a toolkit that provides three key components (Fig \ref{fig:overview}): a Video Synchronizer, a Gaze Target Annotator, and a Video Content Annotator. 

First, to address the common challenge that recordings from different devices cannot always be synchronized at the hardware level during data collection, Video Synchronizer provides a post-hoc synchronization tool that aligns recordings across devices using their audio signals. Second, Gaze Target Annotator enables semi-automatic segmentation of major objects appearing in the scene cameras of head-mounted eye trackers based on a video segmentation model SAM2 \cite{ravi2024sam2segmentimages}. This tool allows both analyzing the statistical distribution of objects in first-person-view videos and mapping gaze location to object categories when eye-tracking data are available. Third, Video Content Annotator automatically annotates activity (here we focus on the bodily states of children and caregivers) in third-person-view videos using a large video–language model (Tarsier 2) \cite{yuan2025tarsier2advancinglargevisionlanguage}. Most of the commercially available artificial intelligence (AI) tools require users to send data to the providers, which may not always be allowed by research ethics protocols. Therefore, we chose to employ open-source AI models that can be deployed on local computers with mid-level modern GPUs.

We demonstrated these three tools on a dataset of child–parent interactions involving children aged 3–4 years (Fig \ref{fig:overview}). In this dataset, both children and caregivers wore head-mounted eye trackers while two fixed cameras in the room record their activities. Together, these tools facilitate efficient and scalable analysis of attention, action, and communication in naturalistic child–caregiver interactions with ego-centric and third-person-view video recordings.

\section{Methods}

\subsection{VIDEO SYNCHRONIZER}
\begin{figure}[t]
    \centering
    \includegraphics[width=\linewidth]{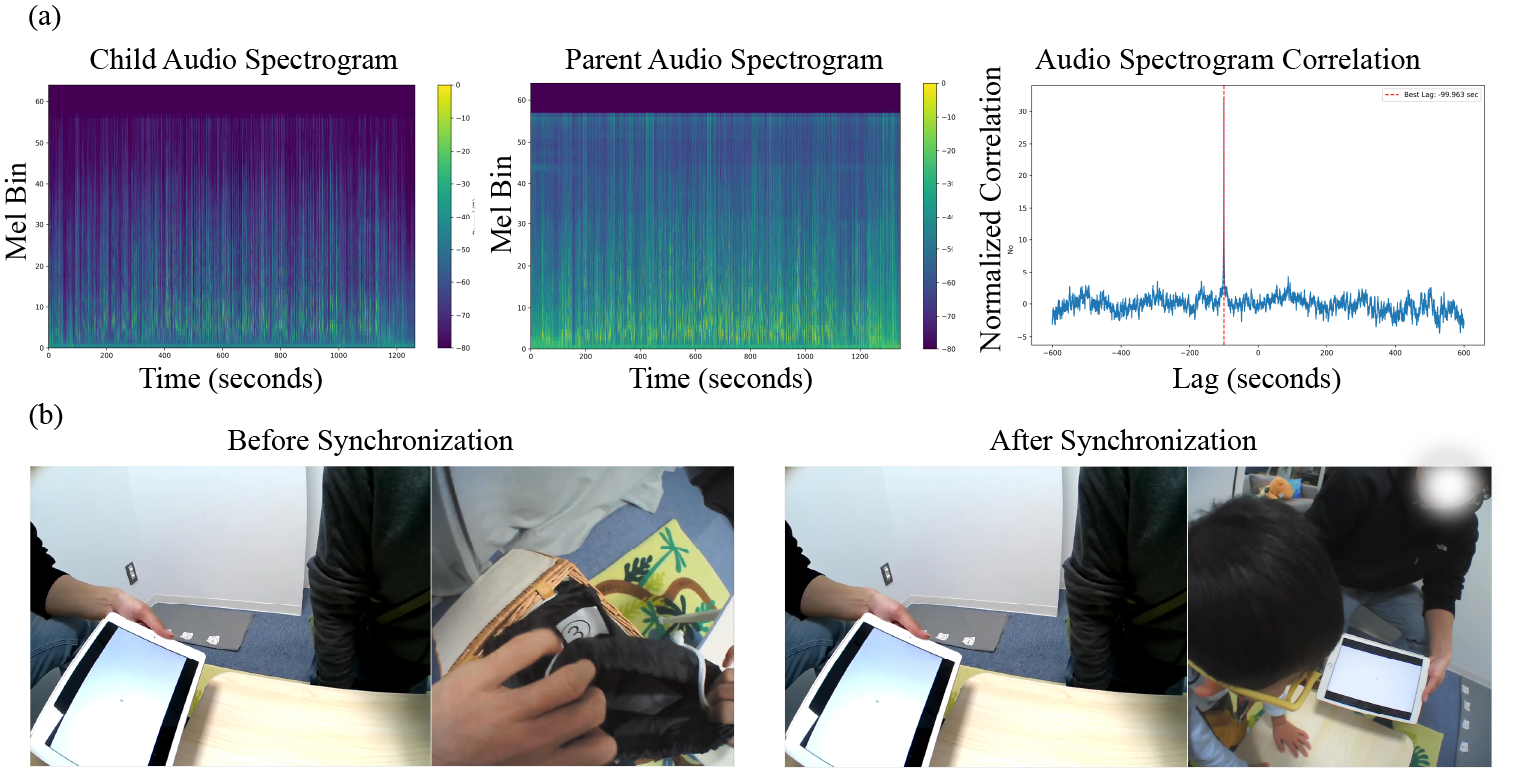}
    \caption{Audio spectrogram–based video synchronizer.
(a) Spectrogram alignment. The two left panels show the audio spectrograms extracted from recordings captured by the child and caregiver cameras, respectively. The right panel shows the normalized cross-correlation between the two spectrograms computed across temporal lags to estimate the optimal synchronization offset.
(b) Synchronization validation. Synchronization performance for an example video is evaluated using a flashing screen stimulus presented during recording. The left panel shows the videos prior to synchronization, where the visual events are temporally misaligned. The right panel shows the videos after synchronization, demonstrating improved temporal alignment between the recordings. }
    \label{fig:synchronizer}
\end{figure}

    To record naturalistic behaviors and social interactions, a multiple camera setup is preferred. However, synchronizing recording devices at the hardware level might be impossible in some cases. Therefore, we develop a post-hoc video synchronizer based on the cross-correlation of audio spectrograms (Fig. \ref{fig:synchronizer}a). A spectrogram represents how acoustic energy is distributed across frequencies over time. To estimate temporal discrepency, we computed the cross-correlation between spectrograms derived from separate audio recordings over a range of temporal lags using 1D convolution. The resulting cross-correlation values were then normalized by a baseline profile obtained from convolving two boxcar time series with the same duration as the spectrograms, which accounts for the correlation expected purely from overlapping signal positions. The lag that gives rise to the maximum correlation is taken as the estimated temporal shift between the two audio streams.
    
    The estimated audio lag is then corrected to synchronize the corresponding video streams of the recording. Following such alignment, all videos could be truncated to the maximal overlapping period to ensure that they have matched endpoints. The synchronizer additionally generates updated time-aligned video frame timestamp and gaze timestamp files for each video, which are required for Gaze Target Annotator.

\subsection{GAZE TARGET ANNOTATOR}
    
    During naturalistic behaviors and social interactions, children and caregivers frequently shift their visual attention across objects and social partners resulting in continuous changes in gaze targets over time. To facilitate efficient annotation of gaze targets in egocentric video recordings with eye-tracking data, we developed a Gaze Target Annotator. The annotator includes two stages: 1. object segmentation based on a prompt-based video segmentation model Segment Anything Model 2 (SAM2) \cite{ravi2024sam2segmentimages} 2. alignment between timestamps of gaze coordinate and that of the video frames to infer gaze targets based on object segmentation masks.
    
    Recent advances in video segmentation models enable semi-automatic object segmentation and tracking in videos with minimal human annotation. In particular, the SAM2 model \cite{ravi2024sam2segmentimages} allows users to indicate the object to segment by annotating a few points on a target object (e.g., a toy) in a few frames. Using these prompts, the model segments the target object in a selected frame and automatically tracks it across the remaining frames, producing segmentation masks that represent the spatial extent of the object throughout the video. To facilitate efficient prompting and annotation, we developed a SAM2-based user interface (UI). The interface allows users to provide positive prompts (points placed within the target object) and negative prompts (points placed on visually similar but irrelevant regions) to guide the segmentation. The UI also includes a frame-wise interactive segmentation module that enables users to test segmentation on a single frame before running inference on the entire video. This step allows users to verify that the prompts are interpreted correctly by the model and adjust them if necessary. Because processing a full video can be computationally intensive, the system supports two workflows: users may either run segmentation directly within the UI or save the annotated prompts and perform segmentation offline using SAM2 while continuing to annotate additional videos. Users can also define custom object lists and select among different model versions, including SAM2 or the newer SAM3 model \cite{carion2025sam}.

    After segmentation is completed, segmentation quality is evaluated using three metrics: Inter-Frame Change (IFC), which quantifies the change in the amount of pixel-level object assignment across consecutive frames; Background Ratio, the proportion of pixels not assigned to any segmented object; and Overlap Score (OS), which measures the degree to which pixels are assigned to multiple objects simultaneously. Users can then refine the segmentation either by adding additional prompts and reprocessing the full video or by selectively correcting problematic sections. Through this iterative process, segmentation quality can be progressively improved until satisfaction.

    With segmentation masks defining the spatial extent of each object in every frame of the egocentric eye-tracking video, the annotator can infer which object is most likely being fixated based on the gaze-point coordinates within each frame. Each gaze sample is modeled as a circular region with a predefined radius around the gaze coordinate to account for eye-tracking noise. For each frame, we compute a confidence ratio: the ratio of pixels in the circular that are assigned to each object. The object with the highest ratio is then assigned as the gaze target for that frame. However, users can also treat the ratios as probabilities of gaze target.

    The original SAM2 model loads all frames of a video on CPU and GPU memory before processing, which prevents it to scale to long videos. We optimized the memory usage by only pre-loading a range of frames beyond the current one being processed to CPU memory and only loading the frame to be processed to GPU memory, which enables the model to process video of arbitrary lengths.  
    
\begin{figure}[t]

    \centering
    \includegraphics[width=\linewidth]{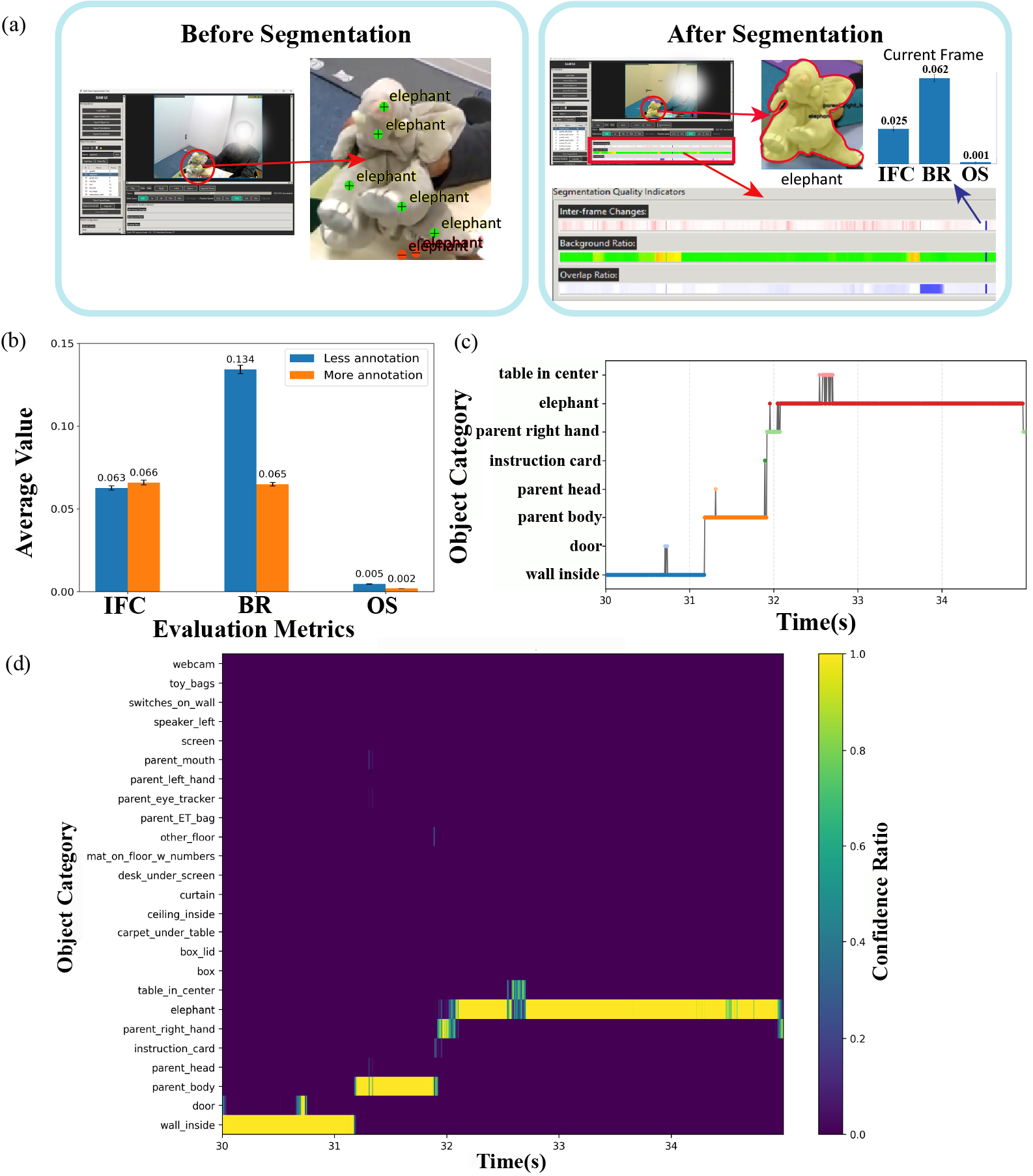}
    \caption{SAM2-based Gaze Target Annotator. (a) Annotation interface. The left panel shows the SAM2-based user interface (UI), which enables users to provide sparse point prompts to guide object segmentation across video frames (e.g., a toy elephant). Positive prompts (colored plus signs) mark the target object, whereas negative prompts (minus signs) suppress assignment of visually or semantically similar objects to the same category. The right panel shows the inspection process, in which segmentation outputs can be reloaded for review. Segmentation masks are overlaid in distinct colors, and the three bars at the bottom display frame-wise segmentation quality scores: inter-frame change ratio (IFC), background ratio (BR), and overlap score (OS). The bar plot shows these scores for the example frame. (b) Temporal average segmentation quality evaluation. The bar plot summarizes the three temporally averaged evaluation scores for a 5-minute child egocentric video example. BR and OS improve as the number of annotation points increases, whereas IFC does not continue to improve because natural object motion introduces unavoidable frame-to-frame variation. (c) Gaze trajectory visualization. A trajectory plot shows the object category label associated with each gaze point over a 5-second segment of the same child video shown in (b). (d) Category confidence visualization. A heatmap shows the confidence ratio of object categories for each gaze event within the same 5-second segment shown in (c). }
    \label{fig:gaze_annotator}
\end{figure}

\subsection{VIDEO CONTENT ANNOTATOR}
Communication is a multi-modal process: language use and eye gaze may vary with the poses and hand actions of both children and caregivers. To investigate such statistics, one needs to have moment-by-moment annotation of the body pose and hand actions of the participants. Recently, large video language models (VLMs) have been developed that allow generating text responses to questions about the content of videos \cite{maaz2024videochatgpt}. We developed a pipeline that utilizes a state-of-the-art open-source model Tarsier 2 (specifically, the variant Tarsier2-7B) to generate time-resolved annotations of video contents. Tarsier 2 encodes individual video frames using a SigLIP visual encoder \cite{zhai2023siglip}, projects visual features through multi-layer perceptron, and reasons about the temporal structure of the video content using a large language model decoder Qwen2 \cite{yang2024qwen2}. 
Our pipeline first extracts overlapping short clips (e.g., 1–3 seconds) from the original video using FFmpeg. Consecutive clips are generated by shifting the time window forward by a fixed interval. This allows adjacent clips to overlap and preserve temporal continuity in the annotations. Each clip is then provided to a VLM together with one or more predefined multiple-choice questions describing the target behaviors or poses to be annotated. The VLM responses are subsequently mapped to a predefined set of categorical labels. These labels are aggregated into an output csv file that records the predicted annotation for each time window. The resulting output file forms a time series of behavioral categories describing the poses or actions of the child and caregiver throughout the video. Each time point corresponds to a short video clip, producing a temporally ordered sequence of predicted behavioral labels. The temporal resolution of this time series is determined by the shift between the onset times of consecutive clips, while the duration of each clip defines the temporal smoothing window over which behaviors are inferred. As users typically need to annotate multiple features (e.g., hand action, body pose, currently engaged toy), we optimized the pipeline by encoding each video clip only once and reusing the keys and values (representations of the content of each patch of the video) for answering all prompts. The user can also arrange prompts sequentially: this allows responses from earlier prompts to provide context for subsequent ones. In practice, we observed that when multiple questions are included within a single prompt, the model occasionally generates answers to unintended questions.

One limitation of vision–language models (VLMs) is that their responses may occasionally exhibit minor variations in wording (e.g., generating “close and facing away from adult” instead of “close but facing away from adult”), even when the temperature is set to its minimum value (deterministically generating
the response with the highest probability). To address this issue, we implemented a cascade of approaches to detect minor anomalies in model outputs and automatically map them to the closest predefined response option. The approaches of capturing anomaly include using a list of customizable aliases of each option to merge them into a unique response (e.g., 'yeah' is an alias of 'yes') and accepting responses of which a pre-defined option is a substring, etc. Outputs that cannot be resolved by this procedure are flagged and presented to the user for manual review. After inspecting the unresolved anomaly, users can identify common variants and update the alias list to further merge response variants into the correct pre-defined response.

\begin{table}[t]
\caption{Example prompts for body state classification}
\label{table_annotation}
\begin{center}
\begin{tabular}{|p{1.8cm}|p{5.5cm}|}
\hline
\textbf{Task} & \textbf{Question} \\
\hline
Child hand action & What is the child's current hand action? Choose one: pointing, grabbing toy, giving away the toy, holding toy still, manipulating toy, gesturing, touching adult, on the ground, on some furniture, resting, none. \\
\hline
Child posture & What is the child's posture? Choose one: sitting still, standing still, walking, crawling, turning around, not visible. \\
\hline
\end{tabular}
\end{center}
\end{table}
    
\subsection{DATASET}

    We evaluated our tools on a dataset of child–caregiver dyadic interactions. In each session, caregivers sat across from their children (aged 3–4 years) and interacted with them while sequentially introducing four toys. The toy set consisted of two familiar animal characters and two unfamiliar Pokémon characters. During each recording session, caregivers presented one toy at a time and described several features of the toy character to the child. The presentation order of the toys was randomized. Each toy was introduced for 4 minutes, after which the caregiver retrieved the next toy from a bag and presented it to the child. Caregivers were provided with instruction cards listing the target features for each toy, but they were encouraged to convey this information using their natural interaction style. Both the caregiver and the child were allowed to physically manipulate the toys during the interaction. Each recording session lasted approximately 20 minutes, and the dataset contains recordings from 41 child–caregiver dyads. The study was approved by The University of Tokyo Office for Life Science Research Ethics and Safety.

    To capture the interaction, both egocentric and third-person-view video recordings were collected. Caregivers wore Pupil Invisible eye-tracking glasses, while children wore the Pupil Neon All Fun and Games eye tracker, which is designed for young children \cite{kassner2014pupil}. Caregiver egocentric videos were recorded at 30 Hz with a frame resolution of 1088 × 1080 pixels and a field of view (FOV) of 82° × 82°. Child egocentric videos were also recorded at 30 Hz, with a frame resolution of 1600 × 1200 pixels and a FOV of 132° × 81°. In addition to egocentric recordings, third-person-view videos were captured using two cameras (Intel realsense D435i and Sony HDR-AS300) positioned in the room: one mounted close to the ceiling and one placed on the side. These fixed cameras were synchronized in recording. They recorded the behaviors of both the caregiver and the child throughout the interaction. At the beginning of each session, a tablet displaying a flashing screen stimulus was presented to both participants and was visible to the side camera. This stimulus provided a visual cue to verify synchronization of recordings across devices.
    
    Audio signals were recorded simultaneously from the caregiver eye tracker, the child eye tracker, and the side camera, with sampling rates of 48 kHz, 44.1 kHz, and 48 kHz, respectively. In addition, gaze data from both eye trackers were recorded at a sampling rate of 200 Hz.





\section{RESULT}
\subsection{Video synchronizer}
To evaluate synchronization accuracy, we manually inspected the temporal alignment of the flashing screen recorded by all cameras across videos using ELAN  \cite{elan2025}. The inspection confirmed that the synchronization algorithm achieved near-perfect alignment, with discrepancies of approximately two frames ($\sim$ 70 ms) across streams (Fig. 2b).

\subsection{GAZE TARGET ANNOTATOR}
    We evaluated segmentation model performance in terms of temporal consistency, segmentation completeness, and segmentation precision using three quantitative metrics: inter-frame change ratio, background ratio, and overlap score.
    
    To assess temporal consistency, we computed the inter-frame change ratio, defined as the proportion of pixels whose segmentation labels change between two consecutive frames. Because objects typically move smoothly across frames, a temporally stable and accurate segmentation should yield a low average inter-frame change ratio. Sudden increases in this metric often correspond to rapid motion of the head or objects, or failures of the model to correctly segment certain objects within a segment.
    
    To evaluate segmentation completeness, we measured the background ratio, defined as the proportion of pixels in a frame that are not assigned any segmentation label. When the object list is complete and the segmentation model performs well, most pixels should be associated with an object mask, resulting in a low background ratio.

    To quantify segmentation precision, we computed the overlap score, which measures the extent to which multiple object masks are assigned to the same physical region due to model confusion. For each pixel, let $k$ denote the number of object masks covering that pixel. The pixel contributes a score of $k-1$ when $k > 1$ and zero otherwise. The frame-level overlap score is obtained by averaging these pixel-wise values across all pixels and normalizing by $ N-1 $, where N is the total number of objects. An ideal segmentation yields an overlap score of 0, indicating that each pixel is assigned to at most one object mask.

    We compared the temporal averages of the three evaluation metrics across segmentations generated with different levels of annotation prompts (Less: on average 5.93 prompts per object on average; More: 22.56). Overall, all three metrics remain close to zero, indicating generally strong segmentation performance across temporal consistency, completeness, and precision. As shown in Fig.~3b, increasing the number of annotation prompts reduces both the background ratio and the overlap score, reflecting improvements in segmentation completeness and precision. In contrast, the inter-frame change ratio remains similar across annotation levels. These results suggest that increasing the number of annotation prompts primarily improves segmentation completeness and precision, and there is a limit in reducing inter-frame change of masks due to the intrinsic motion of objects in videos.
    
    Using the segmentation masks, we align eye-gaze positions with their corresponding object masks to determine the gazed-at target for each tracked gaze sample. The resulting gaze-target trajectories and confidence ratio heatmaps for an example segment (Fig \ref{fig:gaze_annotator}c,d) reveal rapid shifts and variability in gaze targets over time. Additional smoothing may therefore be necessary to obtain more stable gaze-target representations.

    For videos of approximately 20 minutes, 59 categories of objects, the pipeline takes approximately 16 hours to complete on an Nvidia L40s GPU, consuming 3.3 GB of GPU memory. This low memory consumption due to our optimization makes the model feasible to run on most of the consumer-grade GPUs.
    
\subsection{VIDEO CONTENT ANNOTATOR}
\begin{figure}[t]
    \centering
    \includegraphics[width=\linewidth]{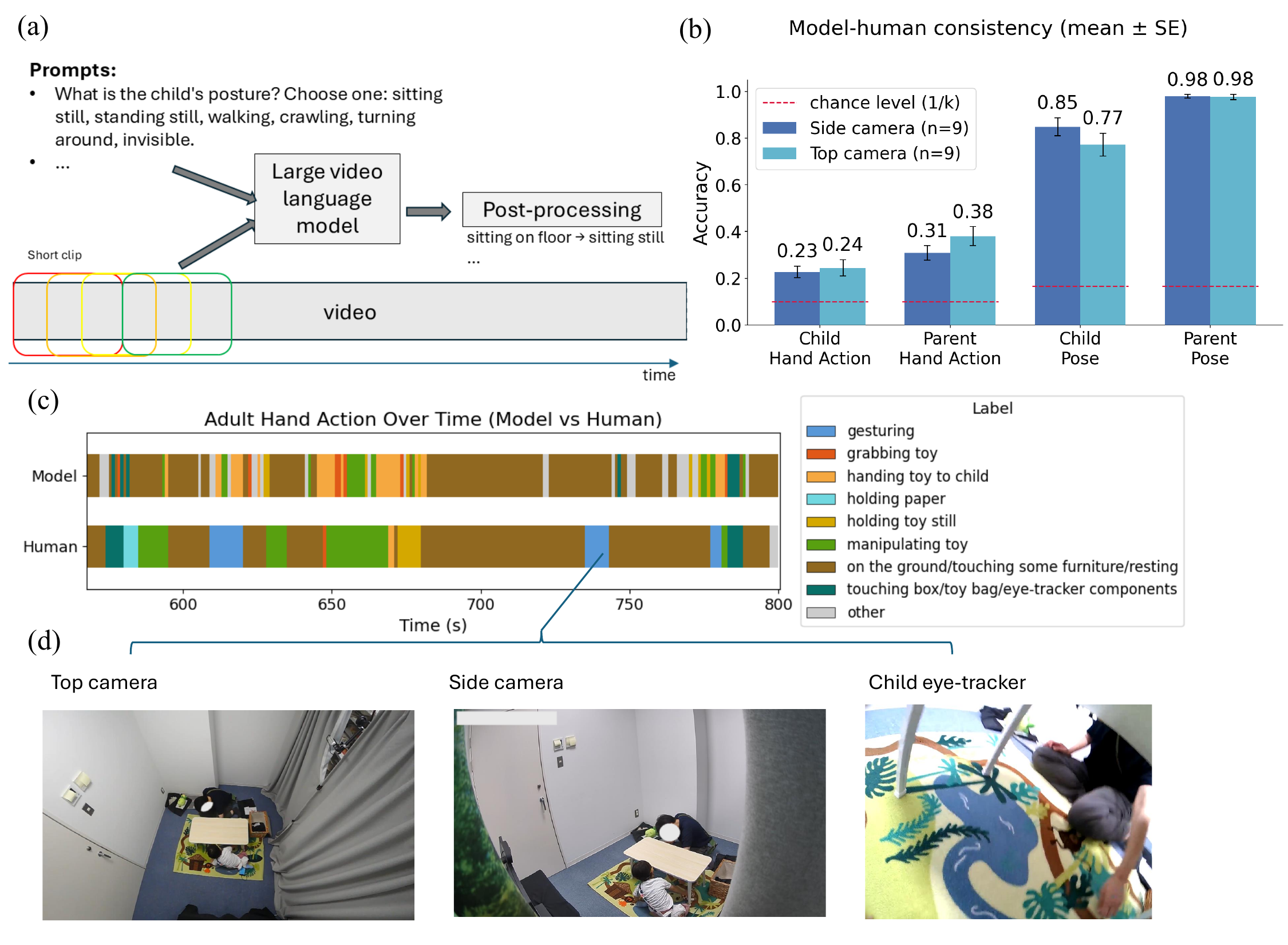}
    \caption{Automatic video content annotator.
(a) Video content annotation. The pipeline employs a video–language model to automatically annotate activities within video clips extracted using sliding temporal windows. The model generates textual descriptions of the observed actions, and then a post-processing step is applied to resolve aliased responses produced by the network.
(b) Annotation agreement between model predictions and human annotations. Example results from four activity segments, each extracted from third-person-view recordings of a different child–caregiver dyad, show that the model achieves the highest agreement with human annotations for caregiver poses, followed by child poses, caregiver hand actions, and child hand actions. (c,d) Failure cases. Errors typically occur under challenging visual conditions, such as when hands are partially occluded from the third-person-view camera or when the activity context is ambiguous. In the example shown, the caregiver was chasing the toy with one hand while occasionally gesturing, and the child was carrying the toy while moving laterally. Because the caregiver’s interacting hand was occluded by the table, the model instead relied on the visible hand resting on the caregiver’s leg to infer the activity.}
    \label{fig:video_annotator}
\end{figure}

We evaluated the consistency between the timeseries output by the automatic pipeline and those produced by human raters. One human annotator annotated the pose and hand action of the caregiver and child in a segment of the recording from each of 9 dyads using ELAN software \cite{elan2025}. 
The annotator can simultaneously view all four videos in each recording and were not aware of the output of the model. The Video Annotator pipeline processed short video clips of 3-second long with 1-second shift of onset between consecutive video clips. A few categories of actions that were functionally similar but treated as separate responses in the prompt to the video annotator pipeline are grouped together as one category for human annotators (e.g., hands on the ground / touching some furniture / resting on the body are all similar as resting or supporting the body). The category of answer that occupied the longest duration in each time window processed by Tarsier 2 was treated as the human annotator's judgment in that time window. Fig \ref{fig:video_annotator}b shows the agreement between the pipeline's output and human annotator on different tasks. Pose estimation achieved higher accuracy than hand action estimation. One possible reason is that full body occupies substantially more pixels in the image. In this dataset, children typically sit facing away from the cameras, which frequently results in their hands being partially or fully occluded in the two camera views. This reduced visibility likely contributes to the lower accuracy observed for hand action estimation. 

Some discrepancies between the model and human annotations might also arise because the model was only applied to third-person camera views, whereas human annotators can examine both egocentric and third-person videos. Figures \ref{fig:video_annotator}c and \ref{fig:video_annotator}d illustrate an example in which the human annotation correctly identified the hand action but the model did not. During the interval shown in Fig. \ref{fig:video_annotator}d, the child moved a toy sideways under the table while the caregiver’s hand followed the toy. This interaction is difficult to infer from the side and top camera views because the table occludes the hands. In contrast, the human annotator can inspect the egocentric eye-tracker recordings and selected the closest category, “gesturing.” In other cases, discrepancies could reflect differences in interpretation rather than annotation errors. For example, one annotator may label a segment as “grabbing toy,” whereas another labels it as “manipulating toy,” and both descriptions can reasonably characterize the observed behavior.

For videos of approximately 20 minutes, processing at 1-second resolution with 3-second time windows (downsampling temporally to 10 fps) for four annotation questions takes approximately 30 minutes, consuming 21.9 GB GPU memory on an Nvidia L40S GPU. Although this experiment used 30 frames from each video clip, using 8 frames is sufficient for many tasks, which requires less GPU memory. Therefore, the pipeline is feasible to be used on local server or workstations when data confidentiality is a requirement.

\section{limitation}
The three tools in GBAT are developed to facilitate analysis of head-mounted eye-tracking data with additional video recording of activities. As demonstated in the result, there are still limitations in their performance. For example, the accuracy of annotating hand actions is much lower than pose. Human annotators perform better because they can integrate both temporal context and views from multiple cameras, while the current pipeline is restrict to processing each video independently, and to a single time window to ensure timing accuracy of the output time courses. One possible solution to overcome the limit of analyzing single video is using Bayesian principle to integrate outputs from multiple videos based on their accuracies, which requires user to provide some amount of ground-truth annotation to estimate accuraices. One solution to provide more temporal context is to increase the time window for each video clip. Another is to include outputs of the model for the past time points as part of the prompt.

The Gaze Target Annotator currently relies on SAM2 model and manual labeling of objects by points. A new SAM3 model \cite{carion2025sam} can accept text as prompt. Future work may integrate both approaches in the video object segmentation part of the Gaze Target Annotator.

\section{CONCLUSIONS}
Head-mounted eye tracking, paired with third-person-view video and audio recordings, provides meaningful data to explore the dynamics underlying children’s and caregivers’ naturalistic behaviors and interactions. To facilitate the annotation of such multimodal, high-dimensional datasets, we developed the GazeBehavior Annotation Toolkit (GBAT). GBAT includes three core components: Video Synchronizer, Gaze Target Annotator, and Video Content Annotator. The Video Synchronizer aligns recordings from multiple devices using audio spectrogram alignment. This aligner enables accurate temporal correspondence across modalities. The Gaze Target Annotator integrates a SAM2-based interface for prompt-guided object segmentation and a pipeline to align annotated object masks with gaze coordinates to determine gaze targets. The Video Content Annotator supports Tarsier 2 based time-resolved labeling of video content, particularly human body states and actions.

Overall, these tools provide an integrated framework for semi-automatic preprocessing and psychologically meaningful annotation of both egocentric and third-person video recordings. To preserve data privacy and ensure practical hardware requirements, the toolkit relies on open-source AI models that can be deployed locally on computers equipped with mid-range modern GPUs. By substantially reducing the manual effort required for large-scale annotation while maintaining high accuracy, GBAT facilitates more efficient and scalable analysis of gaze dynamics and social behaviors in naturalistic caregiver–child interactions.
\addtolength{\textheight}{-3cm}   




\section*{ACKNOWLEDGMENT}
The research was supported by WPI-IRCN startup budget (MBC). We thank Finn Bays for his assistance in annotation. The development of the pipeline utilized Claude Code \cite{anthropic2025claudecode}, OpenAI ChatGPT \cite{achiam2023gpt4}, and some other AI tools.


\bibliographystyle{IEEEtran}
\bibliography{reference}

\end{document}